%% file: root.tex
\documentclass[letterpaper, 10 pt, conference]{ieeeconf}  

\IEEEoverridecommandlockouts                              

\usepackage{graphicx}
\usepackage{lipsum}
\usepackage[linesnumbered,ruled,vlined]{algorithm2e}
\usepackage{amssymb,amsmath,latexsym,amsfonts}
\usepackage[usestackEOL]{stackengine}
\usepackage{amssymb}
\usepackage{caption,url}
\usepackage{float}
\usepackage{epstopdf}
\usepackage{authblk}
\usepackage{dblfloatfix}
\usepackage{multirow}
\usepackage{wrapfig}
\usepackage{cite}

\usepackage{hyperref}
\hypersetup{colorlinks,breaklinks,urlcolor=blue}
\usepackage{dblfloatfix}
\usepackage{multirow}

\usepackage[
  separate-uncertainty = true,
  multi-part-units = repeat
]{siunitx}

\usepackage{subcaption}
\usepackage{gensymb}

\newcommand{\R}{\mathbb{R}}
\newtheorem{theorem}{Theorem}
\newtheorem{lemma}{Lemma}

\usepackage{ulem}

\newcommand{\addred}[1]{\textcolor{red}{\sout{#1}}}

\title{\LARGE \bf
Dynamic Mirror Descent based Model Predictive Control for Accelerating 
Robot Learning 
}

\author{Utkarsh A. Mishra$^{*, 1}$, Soumya R.~Samineni$^{*,1}$, Prakhar Goel$^{2}$, Chandravaran Kunjeti$^{3}$, \\ Himanshu Lodha$^{1}$, Aman Singh$^{1}$, Aditya Sagi$^{1}$, Shalabh Bhatnagar$^{1}$ and Shishir Kolathaya$^{1}$

\thanks{Corresponding author: \url{utkarsh75477@gmail.com}}
\thanks{$^{1}$ Department of Computer Science and Automation, Indian Institute of Science Bangalore}%
\thanks{$^{2}$ Electronics and Communication Engineering Department, Manipal Institute of Technology India}%
\thanks{$^{3}$ Electronics and Communication Engineering Department, National Institute of Technology Karnataka, Surathkal India}%
\thanks{$^{*}$ These authors have contributed equally.}%
}

\begin{document}

\maketitle
\thispagestyle{empty}
\pagestyle{empty}

\begin{abstract}
Recent works in Reinforcement Learning (RL) combine model-free (Mf)-RL algorithms with model-based (Mb)-RL approaches to get the best from both: asymptotic performance of Mf-RL and high sample-efficiency of Mb-RL. Inspired by these works, we propose a hierarchical framework that integrates online learning for the Mb-trajectory optimization with off-policy methods for the Mf-RL. In particular, two loops are proposed, where the Dynamic Mirror Descent based Model Predictive Control (DMD-MPC) is used as the inner loop Mb-RL to obtain an optimal sequence of actions. These actions are in turn used to significantly accelerate the outer loop Mf-RL. We show that our formulation is generic for a broad class of MPC based policies and objectives, and includes some of the well-known Mb-Mf approaches. We finally introduce a new algorithm: Mirror-Descent Model Predictive RL (M-DeMoRL), which uses Cross-Entropy Method (CEM) with elite fractions for the inner loop. Our experiments show faster convergence of the proposed hierarchical approach on benchmark MuJoCo tasks. We also demonstrate hardware training for trajectory tracking in a 2R leg, and hardware transfer for robust walking in a quadruped. We show that the inner-loop Mb-RL significantly decreases the number of training iterations required in the real system, thereby validating the proposed approach.

\end{abstract}



\input{Sections/Introduction.tex}

\input{Sections/Background.tex}

\input{Sections/Methodology.tex}

\input{Sections/Results.tex}
\input{Sections/Hardware.tex}

\input{Sections/Conclusion.tex}

\clearpage

\bibliographystyle{ieeetr}
\bibliography{example}

\clearpage

\input{Sections/Proofs.tex}


\end{document}

%% file: Sections/Introduction.tex
\section{Introduction}


Model-Free Reinforcement Learning (Mf-RL) algorithms are widely applied to solve tasks like dexterous manipulation \cite{Rajeswaran-RSS-18} and agile locomotion \cite{peng2020learning,lee2020learning} as they eliminate the need to model the complex dynamics of the system. However, these techniques are data hungry and require millions of interactions with the environment. Furthermore, these characteristics highly limit successful training on hardware as undergoing such high number of transitions in hardware environments is infeasible. Thus, in order to overcome this hurdle, various works have settled for a two loop model-based approach, typically referred to as Model-based Reinforcement Learning (Mb-RL). Such strategies take the benefit of the explored dynamics of the system by learning the dynamics model, and then determining an optimal policy in this model. Hence this ``inner-loop" optimization allows for a better choice of actions
before interacting  with the original environment. 

\begin{figure}[h]
    \centering
    \includegraphics[width=\linewidth]{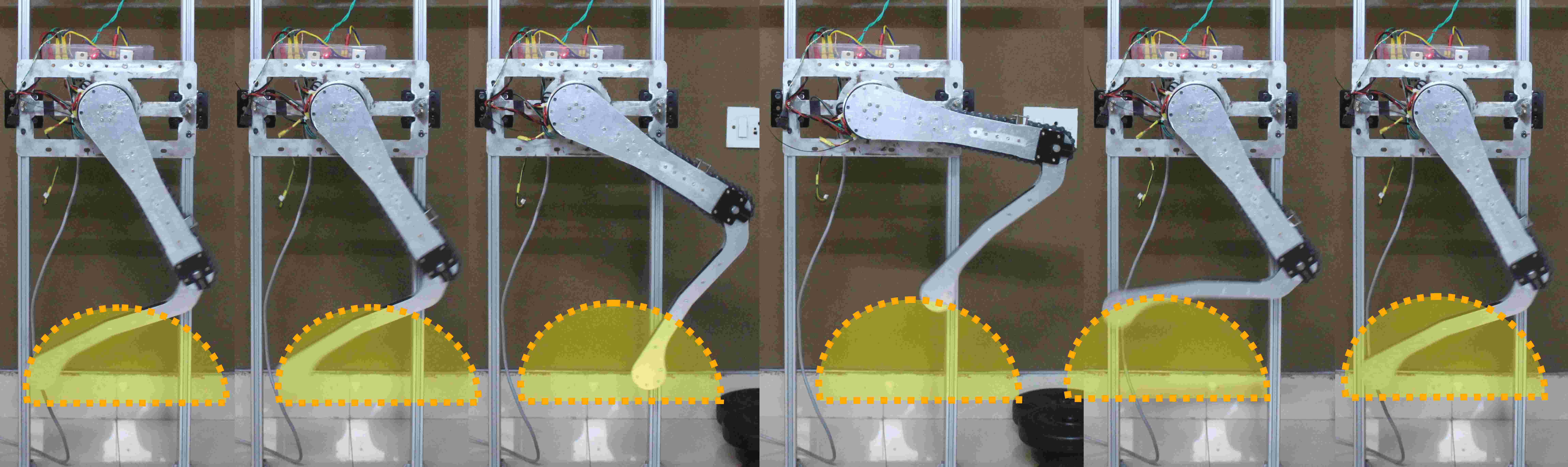}
    \caption{Here the 2R leg is following a semi-elliptical trajectory after successful hardware-in-loop training. The number of training iterations required for the outer-loop hardware RL decreases with the help of inner-loop DMD-MPC. 
    }
    \label{fig:introfigure}
\end{figure}

The inclusion of model-learning in RL has significantly improved sampling efficiency \cite{Levine13GPS,nagabandi2018neural} leading numerous works in this direction.
Such a learned model has proven to be very beneficial in developing robust control strategies (PDDM \cite{nagabandi2020deep} and PETS \cite{NEURIPS2018_handfultrials}) based on predictive simulations. The process of planning with the learnt model is mainly motivated  by the Model Predictive Control (MPC), which is a well known strategy used in classical real-time control. 
Given the model and the cost formulation, a typical MPC structure can be formulated in the form of a finite horizon trajectory optimization problem. With such a view and exploiting the approximated dynamics, methods like 
Cross-Entropy Method (CEM) \cite{pourchot2018cemrl} and Model Predictive Path Integral (MPPI) \cite{Williams2017imppi} have been used to achieve high reward gains. Model Based Policy Optimisation~(MBPO)~\cite{NEURIPS2019_MBPO} introduced the outer loop policy to collect transition to train approximate model and sample over it to train the policy. Gradually, POLO \cite{lowrey2018plan} introduced the use of value functions for terminal rewards in such model based settings, optimizing for which can motivate the policy to converge faster. All these were particularly incorporated by Model Predictive Actor Critic (MoPAC) \cite{morgan2021model} along with optimization in the inner loop using MPPI to accelerate Mb-Mf learning.
Such works demonstrate model-based (Mb) additions to typical model-free (Mf) algorithms accelerating the latter ones with significant sampling efficiency. 

With a view toward strengthening existing Mb-Mf approaches for learning, we propose a generic framework that integrates
a model-based optimization scheme with model-free off-policy learning.
Motivated by the success of online learning algorithms \cite{wagener2019online} in RC buggy models, we combine them with off-policy Mf learning, thereby leading to a two-loop Mb-Mf approach.
%
In particular, we implement dynamic mirror descent (DMD) algorithms on a model-estimate of the system, and then the outer loop Mf-RL is used on the real system. 
The main advantage with this setting is that the inner loop is computationally light; the number of iterations can be large without effecting the overall performance. 
%
%
Since this is a hierarchical approach, the inner loop policy helps improve the outer loop policy, by effectively utilizing the control choices made on the approximate dynamics. This approach, in fact, provides a more generic framework for some of the Mb-Mf approaches (e.g., \cite{morgan2021model}, \cite{nagabandi2020deep}).

In addition to the proposed framework, we introduce a new algorithm Mirror-Descent Model Predictive RL (M-DeMoRL), 
which uses Soft actor-critic (SAC) \cite{haarnoja2018soft} in the outer loop as off-policy RL, and CEM with elite fractions in the inner loop as DMD-MPC \cite{wagener2019online}.
We show that the DMD-MPC accelerates the learning of the outer-loop by simply enriching the off-policy experience data with better choices of state-control transitions. 
    Finally, we demonstrate direct hardware training for 2R leg tracking task (Fig. \ref{fig:legandstochresults}) and hardware transfer of policies for quadruped walking with significantly lesser environment interactions. 





The paper is structured as follows: 
Section \ref{sec:background} will provide the preliminaries for OL for MPC as followed in the paper. Section \ref{sec:methodology} will describe the hierarchical framework for the proposed strategy, followed by the description of the DMD-MPC. Section \ref{sec:result} formulates the algorithm and discusses our simulation results. Section \ref{sec:hardware} presents hardware training on a two-link leg manipulator followed by hardware transfer onto Stochlite quadruped. 
Finally, we 
conclude in Section \ref{sec:conclusion}.

%% file: Sections/Background.tex
\section{Problem Formulation}
\label{sec:background}
We consider an infinite horizon Markov Decision Process (MDP) given by $\{\mathcal{X}, \mathcal{U}, r, P, \gamma, \rho_0 \}$ where $\mathcal{X}~\subset~\mathbb{R}^n$ refers to set of states of the robot and $\mathcal{U}~\subset~\mathbb{R}^m$ refers to the set of control or  actions. $r: \mathcal{X} \times \mathcal{U} \rightarrow \mathbb{R}$ is the reward function, $P : \mathcal{X} \times \mathcal{U} \times \mathcal{X} \rightarrow [0,1]$ refers to the function that gives transition probabilities between two states for a given action, and $\gamma \in (0,1)$ is the discount factor of the MDP. The distribution over initial states is given by $\rho_0: \mathcal{X} \rightarrow [0,1]$ and the policy is represented by $\pi_\theta: \mathcal{X} \to \mathcal{U}$ parameterized by $\theta \in \Theta$, a potentially feasible high-dimensional space. 
If a stochastic policy is used, then $\pi_\theta:\mathcal{X}\times \mathcal{U}\to [0,1]$. 
For ease of notations, we will use a deterministic policy to formulate the problem. Wherever a stochastic policy is used, we will show the extensions explicitly. 
The system model dynamics can be expressed in the form of an equation: $x_{t+1}  \sim f(x_t,u_t)$,
where $f$ is the stochastic transition map.  We can obtain an estimate of this map/model as $f_\phi$, which is parameterized by $\phi$. The goal is to maximise the expected return 
:
\begin{align}\label{eq:infinitehorizon}
    & \theta^* := \arg \max_{\theta} \mathbb{E}_{ \rho_0, \pi_\theta} \left[ \sum^{\infty}_{t = 0} \gamma^t r(x_t, u_t) \right], \\
    & x_0 \sim \rho_0, \quad x_{t+1} \sim f(x_t,\pi_\theta(x_t)).
\end{align}

\textbf{Model Predictive Control (MPC).}
Given the complexity of solving infinite horizon problems \eqref{eq:infinitehorizon} via reinforcement learning (RL), there has been a lot of push toward deployment of Model Predictive Control (MPC) based methods for a finite $H$-step horizon \cite{Levine13GPS}.
Denote the sequence of $H$ states and controls as $\mathbf{x}_t=(x_{t,0}, x_{t,1}, \dots, x_{t,H})$, and $\mathbf{u}_t=(u_{t,0}, u_{t,1}, \dots, u_{t,H-1})$, with $x_{t,0}=x_t$. The cost for $H$ steps is given by
\begin{align}
    C\left(\mathbf{x_t}, \mathbf{u_t}\right) = \sum^{H-1}_{h = 0} \gamma^h c(x_{t,h},u_{t,h}) + \gamma^H c_{H}(x_{t,H}) 
\end{align}
where, $c(x_{t,h}, u_{t,h}) = - r(x_{t,h}, u_{t,h})$ is the cost incurred (for the control problem) and $c_{H}(x_{t,H})$ is the terminal cost\footnote{It will be shown later that in the proposed two-loop scheme, the terminal cost can be the value function obtained from the outer loop.}. Each of the $x_{t,h}, u_{t,h}$ are related by
\begin{equation}
\label{eq:innerloopdynamics}
    x_{t,h+1}  \sim f_\phi(x_{t,h},u_{t,h}), \quad h=0,1,\dots, H-1,
\end{equation}
with $f_\phi$ being the estimate of $f$. We will use the short notation $\mathbf{x_t}\sim f_\phi$ to represent \eqref{eq:innerloopdynamics}.

The solution for \eqref{eq:infinitehorizon} with the finite horizon cost, and with the model estimate $f_\phi$ can be obtained via MPC \cite{Levine13GPS,nagabandi2018neural,NEURIPS2019_deepdynamicmodel}. At every step $t$, optimal sequence of actions/controls are obtained. The first action is then applied on the real system to obtain the next state. 
There are several ways to solve the MPC setup, and online learning (OL) is one such approach, which is described next.

\textbf{Online Learning for MPC.}
Online Learning (OL) is a generic sequential decision making technique that makes a decision at time $t$ to optimise for the regret over time. 
Since MPC also involves taking optimal decisions sequentially, \cite{wagener2019online} proposed to use online learning via dynamic mirror descent (DMD) algorithms.
DMD is reminiscent of the proximal update with a Bregman divergence that acts as a regularization to keep the current control distribution 
close to the previous one. 
For a rollout time of $H$, we sample the tuple $\mathbf{u_t}$ from a control distribution ($\pi_\eta$) parameterized by $\eta \in \mathcal{P}$, where $\mathcal{P}$ is the parameter set. To be more precise, $\eta_t$ is also a sequence of parameters: $\eta_t= (\eta_{t,0},\eta_{t,1}, \dots, \eta_{t,H-1})$ which yield the control tuple $\mathbf{u_t}$. Therefore, given the control distribution paramater $ \eta_{t-1}$ at round $t-1$,
we obtain $\eta_t$ at round $t$ from the following update rule:
\begin{align}
\label{eq:costmin}
    & J(x_t,\tilde \eta_t) \! := \! \mathbb{E}_{\mathbf{u_t}\sim \pi_{\tilde \eta_t},  \mathbf{x_t}\sim f_\phi} \left [ C(\mathbf{x_t},\mathbf{u_t}) \right], \! \tilde{\eta}_{t}\! :=\! \Phi_t (\eta_{t-1})  \\
    \label{eq:3}
& \eta_t = \arg \min_{\eta} \ \left [ \alpha_t  \langle \nabla_{\tilde \eta_t} J(x_t,\tilde \eta_t) , \eta  \rangle + D_{\psi} (\eta \| \tilde \eta_t)  \right ] , 
\end{align}
where 
$J$ is the MPC objective/cost expressed in terms of $x_t$ and $\pi_{\tilde \eta_t}$, 
$\Phi_t$ is the shift model, $\alpha_t >0$ is the step size for the DMD, and $D_{\psi}$ is the Bregman divergence for a strictly convex function $\psi$. Note that the shift parameter $\Phi_t$ is critical for convergence of this iterative procedure. Typically, this is ensured by making it dependent on the state $x_t$. In particular, for the proposed two-loop scheme, we make $\Phi_t$ dependent on the outer loop policy $\pi_{\theta}(x_t)$. 
Also note that resulting parameter $\eta_t$ is still state-dependent, as the MPC objective $J$ is dependent on ${x_t}$.

\begin{figure*}[t]
    \centering
    \includegraphics[width=\linewidth]{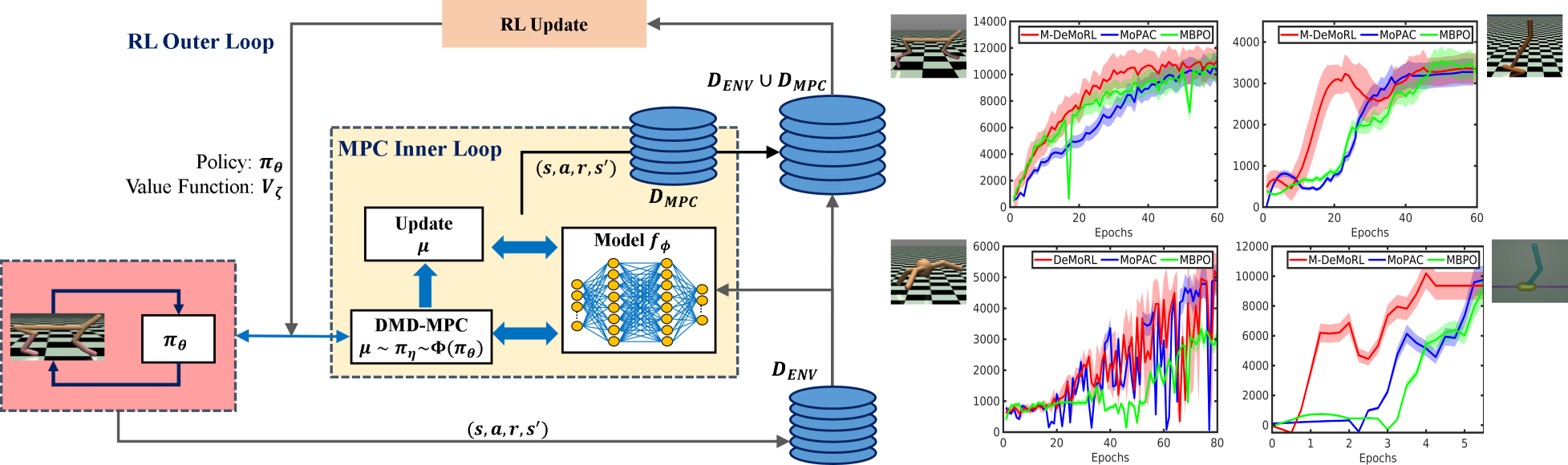}
    \caption{ (Left) The proposed hierarchical structure of Mirror-Descent Model Predictive Reinforcement Learning (M-DeMoRL) with an inner loop DMD-MPC update and an outer loop RL update. (Right) The Mujoco OpenAI benchmark environments solved by the proposed algorithm with corresponding performance plots compared with other model based algorithms: MoPAC and MBPO.}
    \label{fig:method}
\end{figure*}


With the two policies, $\pi_{\theta}$ and $\pi_{\eta_t}$ at time $t$, we aim to develop a synergy in order to leverage the learning capabilities of both of them. In particular, the ultimate goal is to learn them in ``parallel", i.e., in the form of two loops. The outer loop optimizes $\pi_{
\theta}$ and the inner loop optimizes $\pi_{\eta_t}$ for the MPC Objective. 

%% file: Sections/Methodology.tex
\section{Methodology}
\label{sec:methodology}

In this section, we discuss a generic approach for combining model-free (Mf) and model-based (Mb) reinforcement learning (RL) algorithms through DMD-MPC. In classical Mf-RL, data from the interactions with the original environment are used to obtain the optimal policy parameterized by $\theta$. 
While the interactions of the policy are stored in memory buffer, $\mathcal{D}_{ENV}$, for offline batch updates, they are used to optimize the parameters $\phi$ for the approximated dynamics of the model, ${f}_\phi$. Such an optimized policy can then be used in the DMD-MPC strategy to update the control distribution, $\pi_{\eta}$. The controls sampled from this distribution are rolled out with the model, ${f}_\phi$, to collect new transitions and store these in a separate buffer $\mathcal{D}_{MPC}$. Finally, we 
%
update $\theta$ 
using both the data i.e., from the buffer $\mathcal{D}_{ENV} \cup \mathcal{D}_{MPC}$ 
via one of the off-policy approaches (e.g. DDPG \cite{lillicrap2015continuous}, SAC \cite{haarnoja2018soft}). 
In this work, we majorly demonstrate this using Soft Actor-Critic (SAC) \cite{haarnoja2018soft}. 
This gives a generalised hierarchical framework with two loops: 
DMD-MPC forming an inner loop and model-free RL in the outer loop. A graphical representation of the described framework is given in Figure~\ref{fig:method}.

There are two salient features in the two-loop approach:
\begin{itemize}
    \item[1.] At round $t$, we obtain the shifting operator $\Phi_t$ by using the outer loop parameter $\theta$. This is in stark contrast to the classical DMD-MPC method shown in \cite{wagener2019online}, wherein the shifting operator is only dependent on the control parameter of the previous round $\eta_{t-1}$.
    
\item[2.] Inspired by \cite{lowrey2018plan,morgan2021model}, the terminal cost $c_{H}(x_{t,H})=-~V_\zeta(x_{t,H})$ is the value of the terminal state for the finite horizon problem as estimated by the value function ($V_\zeta$, parameterized by $\zeta$) associated with the outer loop policy, $\pi_{\theta}$. This will efficiently utilise the model learned via the RL interactions and will in turn optimize $\pi_{\theta}$ with the updated setup. 
\end{itemize}

Since there is limited literature on theoretical guarantees of DRL algorithms, it is difficult to show convergences and regret bounds for the proposed two-loop approach. 
However, there are guarantees on regret bounds for DMD algorithms in the context of online learning \cite{hall2013dynamical}.
We reuse their following definitions:
\begin{gather*}
G_J \triangleq \max_{\eta_t \in \mathcal{P}} \|\nabla J(\eta_t)\|, \quad M \triangleq \frac{1}{2} \max_{\eta_t \in \mathcal{P}} \|\nabla \psi(\eta_t)\| \\
D_{max} \triangleq \max_{\eta_t,\eta_t' \in \mathcal{P}} D_\psi(\eta_t\|\eta_t'), \\
\Delta_{\Phi_t} \triangleq \max_{\eta_t,\eta_t' \in \mathcal{P}} D_\psi(\tilde\eta_t\|\tilde\eta_t') - D_\psi(\eta_t\|\eta_t').
\end{gather*}
By a slight abuse of notations, we have omitted $x_t$ in the arguments for $J$. We have the following:

\ifdefined\reviewON
\begin{theorem}
\label{theo1}\addred{
    Given the shift operator $\Phi_{t}$ that is dependent on the outer-loop policy parameterised by $\theta$ at state $x_t$, the Dynamic Mirror Descent (DMD) algorithm using a diminishing step sequences $\alpha_{t}$ gives the overall regret with the comparator sequence $\eta_t$ as,}
\begin{equation}
\addred{\mathcal{R}\left({\eta}_{T}\right) = \sum_{t=0}^{T} J\left(\tilde{\eta}_{t}\right)-J\left(\eta_{t}\right) \leq \frac{D_{\max }}{\alpha_{T+1}}+\frac{4 M}{\alpha_{T}} W_{\Phi_{t}}\left({\eta}_{T}\right)+\frac{G_{J}^{2}}{2 \sigma} \sum_{t=0}^{T} \alpha_{t}}
\end{equation}\addred{
with}
\begin{equation*}\addred{
W_{\Phi_{t}}\left({\eta}_{T}\right) \triangleq \sum_{t=0}^{T}\left\|\eta_{t+1}-\Phi_t(\eta_{t})\right\|.}
\end{equation*}\addred{
Based on such a formulation, the regret bound is
$\mathcal{R}\left({\eta}_{T}\right)=O\left(\sqrt{T}\left[1+W_{\Phi_{t}}\left({\eta}_{T}\right)\right]\right)$.}
\end{theorem}
 \addred{  Proofs of both Lemma \ref{lem1} and Theorem \ref{theo1} are given in \cite{hall2013dynamical}. Theorem \ref{theo1} shows that the regret is bounded by $\|\eta_{t+1}-\Phi_t(\eta_{t})\|$, where the shifting operator $\Phi_t$ is dependent on the outer-loop policy.
With the shift model obtained from RL policy at every iteration, the regret decreases as the policy converges and better approximates the comparator sequence $\eta_{t+1}$ using the learnt dynamics by forwarding system dynamics.
However, this result is not guaranteed for non-convex objectives, which will be a subject of future work. 
}\fi

\begin{lemma}
\label{nlem1}
Denote the optimal cost obtained for \eqref{eq:costmin} for the real model by $J_r$. Then the difference in the returns between the real and the approximated model ($J_r$ and $J$ respectively) is 
\begin{align*}
     {J}\left(x_t,\tilde{\eta}_{t}\right)-J_r\left(x_t,\tilde{\eta}_{t}\right) \! \leq &  2  c_{max} \frac{(H\!-\!1)\gamma^{H+1}\! - \! H\gamma^H  + \gamma}{(1-\gamma)^2}  \epsilon_f \\
     & + \gamma^H 2 \ V_{max} H \epsilon_f \triangleq R_{f,H}
\end{align*}
using Lemma B.3 in \cite{NEURIPS2019_MBPO}. Here, $\epsilon_f$ is the uncertainty in dynamics model approximation. 
\end{lemma}

\begin{theorem}
\label{thm:regrettheorem}
Let the  $\tilde{\eta}_{t}$ be sequence obtained from outer loop policy $\pi_{\theta}$, and let ${\eta}_{t}^\star$ be the policy obtained from $\pi_{\theta^\star}$, where ${\theta^\star}$ is optimal;
then for the class of convex MPC objectives $J$, the maximum regret incurred for the $T$ decision timesteps, each with $H$-step planning  can be formulated as 
\begin{align*}
     {Re}_{T} \left(\boldsymbol{\eta}_{T}\right) &  := \sum_{t=0}^T {J}\left(\tilde{\eta}_{t}\right)-J_r\left({\eta}_{t}^{\star}\right)  \\
     & \leq \! \frac{D_{\max }}{\alpha_{T+1}} \! + \! \frac{4 M}{\alpha_{T}} W_{\Phi_{t}}\left(\boldsymbol{\eta}_{T}\right)+\frac{G_{J}^{2}}{2 \sigma} \sum_{t=1}^{T} \alpha_{t} + T \ R_{f,H} 
\end{align*}
where $\boldsymbol{\eta}_{T}$ is the vector of parameters decided at every decision step, $W_{\Phi_t}$ is given by
\begin{equation}
    W_{\Phi_{t}} = \sum_{t=0}^T \| \eta_{t+1}^{\star}-\tilde\eta_{t+1}\|,
\end{equation} and other notations are derived from \cite{hall2013dynamical}.
\end{theorem}
Proofs of Lemma \ref{nlem1} and Theorem \ref{thm:regrettheorem} are provided in Appendix.

As  we  obtain  the  shift  model  at  every  iteration  using the outer loop RL policy and the learned dynamics,  the  maximum regret  decreases  as  the  policy ${\theta}$ converges to ${\theta^\star}$, the sequence $\tilde\eta_{t+1}$ approaches $\eta_{t+1}^{\star}$. In other words, the regret is minimum when the infinite horizon optimal outer loop policy is efficient enough in identifying a finite H-step horizon optimal inner loop policy.


\textbf{DMD-MPC with Exponential family.}
We consider a parametric set of probability distributions for our 
control distributions in the exponential family, given by natural parameters $\eta$, sufficient statistics $\delta$ and expectation parameters $\mu$ \cite{wagener2019online}.
Further, we set Bregmann divergence in \eqref{eq:3} to the KL divergence. 
The natural parameter of control distribution, $\tilde{\eta_t}$, is obtained with the proposed shift model $\Phi_t$ from the outer loop RL policy $\pi_{\theta}$
by setting the expectation parameter of $\tilde\eta_t$: $\mathbf{\tilde \mu_t}=  \pi_{\theta}(\mathbf{x_t})$.  Note that we have overloaded the notation $\pi_{\theta}$ to map the sequence $\mathbf{x_t}$ to $\mathbf{\tilde \mu_t}$, which is the sequence of $\tilde \mu_{t,h} = \pi_{\theta}\left(x_{t,h}\right)$\footnote{Note that if the policy is stochastic, then $ \tilde{\mu}_{t,h} \sim \pi_{\theta}\left(x_{t,h}\right)$. This is similar to the control choices made in \cite[Algorithm 2, Line 4]{morgan2021model}.}.
Then, we have the following gradient of the cost:
\begin{align}
    \nabla_{\tilde \eta_t} J(\mathbf{x_t}, \tilde \eta_t) = \mathbb{E}_{\mathbf{u_t} \sim \pi_{\tilde \eta_t},\mathbf{x_t} \sim f_\phi} \left [ C(\mathbf{x}_t, \mathbf{u}_t)(\delta(\mathbf{u}_t) - \mathbf{\tilde{\mu}_t} )  \right ],
\end{align}
 In the presented setup, we choose Gaussian distribution for control and $\delta(\mathbf{u_t}) := \mathbf{u_t}$. We finally have the following update rule for the expectation parameter \cite{wagener2019online}:
\begin{equation}
\label{eq:dmd-mpcexponentialfamily}
    \mathbf{\mu_t}=\left(1-\alpha\right) \mathbf{\tilde\mu_t}+\alpha \mathbb{E}_{\pi_{\tilde \eta_t}, f_\phi}\left[C\left(\mathbf{x_{t}}, \mathbf{u_t}\right) \mathbf{u_t}\right].
\end{equation}

Based on the data collected in the outer loop, the inner loop is executed via DMD-MPC as follows:
\begin{itemize}
\item Step 1. 
Considering $H$-step horizon, for $h=0,1,2, \dots, H-1$, obtain 
\begin{align}
\label{eq:innerDMDMPC1}
& \tilde \eta_{t,h}= \Sigma^{-1} \tilde \mu_{t,h}, \quad \tilde \mu_{t,h}=\pi_{\theta}(x_{t,h}) \\
\label{eq:innerDMDMPC2}
& u_{t,h} \sim \pi_{\tilde{\eta}_{t,h}}  \\
\label{eq:innerDMDMPC3}
& x_{t,h+1} \sim f_\phi(x_{t,h}, u_{t,h}). 
\end{align}
where 
$\Sigma$ represents the 
covariance for control distribution.
    \item Step 2. Collect 
        $\tilde \eta_t = ( \tilde \eta_{t,0}, \tilde \eta_{t,1}, \dots, \tilde \eta_{t,H-1}
    )$,
and apply DMD-MPC \eqref{eq:3} to obtain $\eta_t$.
\end{itemize}

For the presented work, we use CEM with the method of elite fractions that allows us to select only the best transitions. This is given by the following:
\begin{align}
\label{eq:cemobjective}
J(\mathbf{x_t},\tilde \eta_t) := -\log \mathbb{E}_{\pi_{\tilde \eta_t}, f_\phi} \left[\boldsymbol{1}\left\{ C\left(\mathbf{x_t}, \mathbf{u_t}\right) \leq C_{t, \max }\right\}\right]
\end{align}
where we choose $ C_{t,max}$ as the top elite fraction from the estimates of rollouts. It is worth noting that both CEM and MPPI belong to the same family of objective function utilities \cite{wagener2019online}.




%% file: Sections/Results.tex
\section{Implementation and Results}
\label{sec:result}

In this section, we implement the two-loop hierarchical framework as explained in the previous section and structure the specific details about the algorithm associated with the proposed work. This will be compared with the existing approaches MoPAC \cite{morgan2021model} and MBPO \cite{NEURIPS2019_MBPO} on the benchmark MuJoCo control environments.

\subsection{Algorithm: 
}

M-DeMoRL algorithm derives from other Mb-Mf methods in terms of learning dynamics and follows a similar ensemble dynamics model approach. We have shown it in Algorithm \ref{algorithm}. There are three parts in this algorithm: Model learning, Soft Actor-Critic and DMD-MPC. We describe them below.

\begin{algorithm}[t]
    \caption{M-DeMoRL Algorithm}
    \SetAlgoLined
    \label{algorithm}
    $ \text {Initialize SAC and Model } f_\phi, \text{Environment Parameters} $\\
    $ \text {Initialize memory buffer: } D_{ENV}$ \\ 
    \For{max iterations}{
    $D_{ENV} \leftarrow D_{ENV} \cup\left\{x, u, r, x'\right\}, u \sim \pi_{\theta}\left(x\right)$\\
    \For {each model learning epoch}{
    $\text { Train model } f_{\phi} \text{ on } D_{ENV} \text{ with loss}: J_{\phi} = \|\{(x' - x), r\} - f_\phi(x,u)\|_2$}
    
    Initialize $D_{MPC}$\\
    Calculate M and H from schedule \\
    \For{ \text{DMD-MPC iterations}}{
    $\text {Sample } x_{t,0} \text { uniformly from } D_{ENV} $\\
   $\text{Simulate M trajectories of H steps horizon: }$ \eqref{eq:innerDMDMPC1}, \eqref{eq:innerDMDMPC2} and \eqref{eq:innerDMDMPC3}\\
    $\text{Perform CEM to get optimal action sequence: } \mathbf{\mu_t}$ \eqref{eq:elitecalc1} and \eqref{eq:dmd-mpcexponentialfamily}\\
    $\text{Collect complete trajectory: } \mathbf{x_t}, \mathbf{r_t} \sim f_\phi(x_{t,0}, \mathbf{\mu_t})$ \\
    $ \text {Add all transitions to } D_{MPC}: D_{MPC} \leftarrow D_{MPC} \cup\left\{x_{t,h}, u_{t,h}, \hat{r}_{t,h}, x_{t,h+1}\right\}$   
    }
    \For {each gradient update step} {
    Update SAC parameters using data from $D_{ENV} \cup D_{MPC}$
    }
    }
\end{algorithm}

\textbf{Model learning.} The functions to approximate the dynamics and reward function of the system are $K$-probabilistic deep neural networks \cite{NEURIPS2019_deepdynamicmodel} 
 cumulatively represented as 
$\{f_{\phi_1}, f_{\phi_2}, \dots, f_{\phi_K}\}$
Using the inputs as the current state and actions, the ensemble model fits all the probabilistic models to output change in states and the reward obtained during the transition. 
Such a configuration is believed to account for the epistemic uncertainty of complex dynamics and overcomes the problem of over-fitting generally encountered by using single models~\cite{NEURIPS2018_handfultrials}. 

\textbf{SAC.} Our implementation of the proposed algorithm uses Soft Actor-Critic (SAC) \cite{haarnoja2018soft} as the model-free RL counterpart. Based on principle of entropy maximization, the choice of SAC ensures sufficient exploration motivated by the soft-policy updates, resulting in a good approximation of the underlying dynamics. 

\textbf{DMD-MPC.} Here, we solve for $\mathbb{E}_{\pi_{\tilde \eta_t}, f_\phi}\left[C\left(\mathbf{x_{t}}, \mathbf{u_{t}}\right) \mathbf{u_{t}}\right]$ using a Monte-Carlo estimation approach. For a horizon length of $H$, we collect $M$ trajectories using the current policy $\pi_{\theta_t}$ and the more accurate dynamic models from the ensemble having lesser validation losses. For all trajectories, the complete cost is calculated using a deterministic reward estimate and the value function through (2). After getting the complete state-action-reward $H$-step trajectories we execute the following based on the CEM \cite{pourchot2018cemrl} strategy:
\begin{itemize}
    \item Step 1. Choose the $p\%$ elite trajectories according to the total $H$-step cost incurred. We set $p~=~10~\%$ for our experiments, and denote the chosen respective action trajectories and costs as $U_{elites}$ and $C_{elites}$ respectively. 
    Note that we have also tested for other values of $p$, and the ablations are shown later in this section.
    
    \item Step 2. Using $U_{elites}$ and $C_{elites}$ we calculate $\mathbb{E}_{\pi_{\tilde \eta_t}, f_\phi}\left[C\left(\mathbf{x_{t}}, \mathbf{u_{t}}\right) \mathbf{u_{t}}\right]$ as the reward weighted mean of the actions i.e.
    \begin{align}
    \label{eq:elitecalc1}
        \mathbf{g_t} = \frac{\sum_{i \in elites} {C}_{i} \ U_{i}}{\sum_{i \in elites} {C}_{i}}
    \end{align}
    and update the current policy actions, $\mathbf{\tilde\mu_t} = \pi_{\theta_i}(\mathbf{x_t})$ according to \eqref{eq:dmd-mpcexponentialfamily}.
    \ifdefined \reviewON \addred{
    \item[Step 3] Finally, we update the current policy actions, $\mathbf{\tilde\mu_t} = \pi_{\theta_t}(\mathbf{x_t})$ according to \eqref{eq:dmd-mpcexponentialfamily} as
    \begin{align}
    \label{eq:elitecalc2}
        \mathbf{\mu_{t}}=\left(1-\alpha\right) \mathbf{\tilde\mu_t} +\alpha \mathbf{g_t}.
    \end{align}}\fi
\end{itemize}

\subsection{Experiments and Comparison}

Several experiments were conducted on the MuJoCo \cite{todorov2012mujoco} continuous control tasks with the OpenAI-Gym benchmark and the performance was compared with recent related works MoPAC \cite{morgan2021model} and MBPO \cite{NEURIPS2019_MBPO}. First, we discuss the hyperparameters used for all our experiments and then the performance achieved in the conducted experiments.

As the baseline of our framework is built upon MBPO implementation, we derive the ``same hyperparameters" for our experiments and all the baseline algorithms. We compare the results of three different seeds and the reward performance plots are shown in Figure \ref{fig:method}(right). 

For the inner DMD-MPC loop we choose a varying horizon length from $5-15$ and perform $100$ trajectory rollouts. With our elite fraction as $10\%$, the updated model-based transitions are added to the MPC-buffer. This process is iterated with a varying batch-size with maximum of $10,000$ thus completing the DMD-MPC block in Algorithm \ref{algorithm}. These variable batch size and horizon length allows us to exploit the models more when we have achieved significant learning considering uncertainties and distribution shifts. Also, as evident from Theorem~1, increasing H directly increases the maximum regret incurred. 
Thus, we start from low horizon length as the regret incurred is more in the initial phases and gradually increase the horizon length to exploit the capabilities of MPC. Further, the number of interactions with the true environment for outer loop policy were kept constant to $1000$ for each epoch, same as MoPAC and MBPO. 

We clearly note an accelerated progress for all the environments, with approximately $30\%$ faster rate in the reward performance curve as compared to best of prior works. Our rewards in Ant-v2 were comparable with MoPAC but still significantly better than MBPO. We would like to emphasize that our final rewards are eventually the same as achieved by MoPAC and MBPO, however the progress rate is faster for all our experiments with lesser true environment interactions. 

\subsection{Ablation study on elite percentage}

Given the sequence of controls $\mathbf{\mu_t}$, we collect the resulting trajectory and add them to our buffer. Therefore, the quality of $\mathbf{\mu_t}$ is a significant factor affecting the quality of data used for the outer loop RL-policy. The selection strategy being CEM, a quality metric is dependent on the choice of elite fractions $p$. 
\begin{figure}[t]
    \centering
    \includegraphics[width=\linewidth]{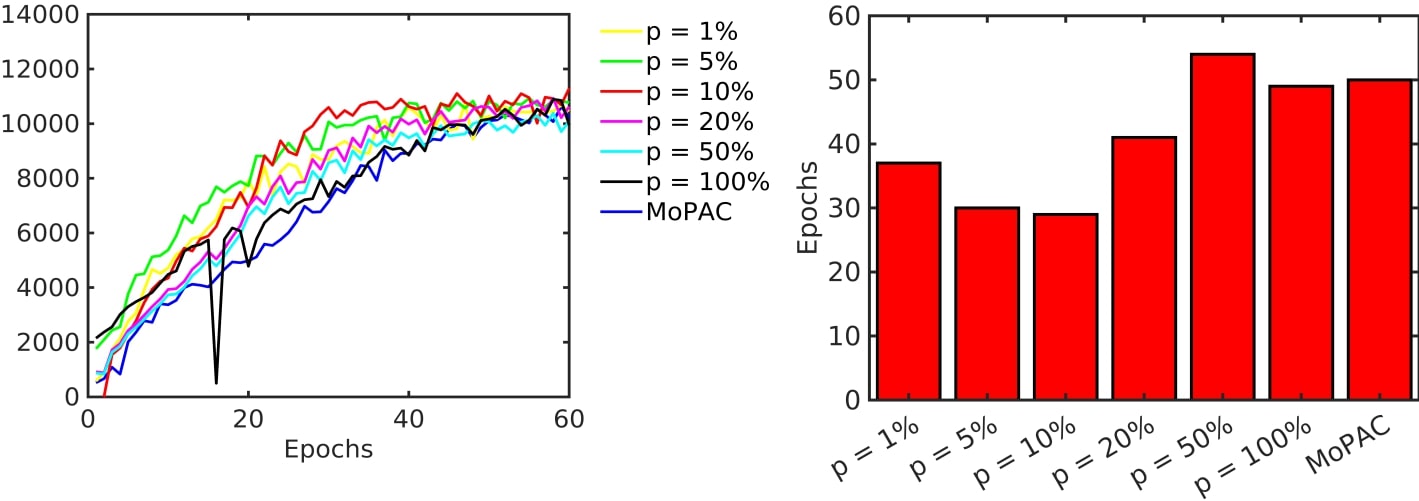}
    \caption{Ablation study for elite percentage: Reward performance curve (left) and Acceleration analysis as epochs to reach 10000 rewards (right)}
    \label{fig:ablation}
\end{figure}

\begin{figure*}[t]
    \centering
    \includegraphics[width=\linewidth]{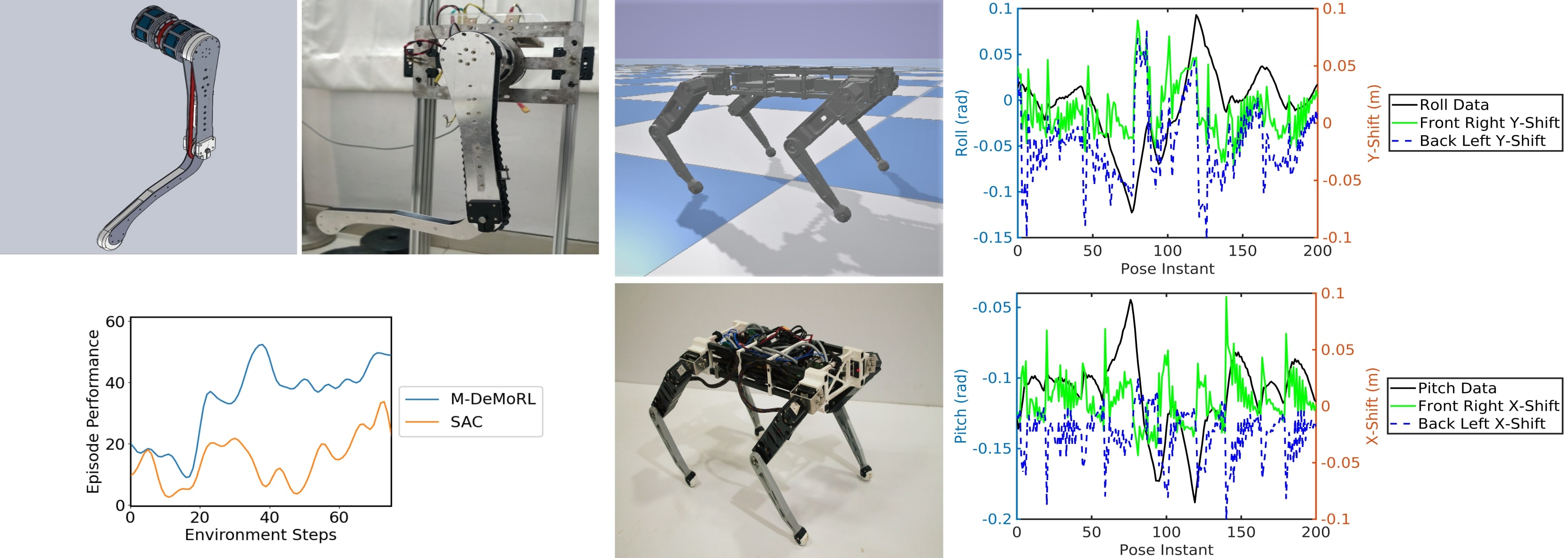}
    \caption{(Left) Two link Leg Manipulator and corresponding hardware test setup followed by Stochlite quadruped Model Design, Pybullet Simulation and Hardware as incorporated for hardware transfer. The hardware-in-loop training for the leg shows faster reward gains compared SAC. Similarly, the transfered policy in StochLite shows desirable variations in the X and Y shifts based on the torso orientation. 
    }
    \label{fig:legandstochresults}
\end{figure*}

We perform an ablation study for $6$ values of $p~=~1, 5, 10, 20, 50$ and $100\%$ on HalfCheetah-v2 OpenAI gym environment. The analysis was performed based on the reward performance curves as shown in Fig.~\ref{fig:ablation} (left). Additionally, we realize the number of the epochs required to reach a certain level of performance as a good metric to measure acceleration achieved. Such an analysis is provided in Fig.~\ref{fig:ablation} (right). We make the following observations:
\begin{itemize}
    \item Having a lesser value of $p$ might ensure that learned dynamics is exploited the most, but decreases the exploration performed in the approximated environment.
    \item Similarly, having higher value of $p$ on the other hand will do more exploration using a ``not-so-perfect" policy and dynamics. 
\end{itemize}
Thus, the elite fraction balances between exploration and exploitation. 

%% file: Sections/Hardware.tex
\section{Hardware training and transfer}
\label{sec:hardware}


In this section, we extend M-DeMoRL to more complex dynamical systems like 2R quadruped leg and a complete quadruped walking task as well. In the first case, we do direct algorithm transfer to execute learning with hardware, whereas for quadruped we show generation of hardware transferable policies.

\textbf{Leg trajectory tracking:} The main idea here is to track a fixed end-foot trajectory with a 2R quadruped robot leg (as shown in Figure~\ref{fig:legandstochresults} left). The leg is torque controlled and equipped with hall effect sensors and BLDC motors. In order to perform direct hardware training, we accomplish torque control at 500 Hz, with the motor controller running at 40KHz. We define the task as tracking an elliptical trajectory (as shown in Figure~\ref{fig:legandstochresults}. The reward is chosen as follows:
\begin{align}
    r_{leg} := 0.8 e^{-\|p-p_d\|^2} + 0.2 e^{-\|\dot p - \dot p_d\|^2},
\end{align}
where $p,p_d\in \R^2$ are the actual and desired end-effector positions of the leg.
We observed enhanced learning performance as compared to baseline SAC within 20 epochs (20k interactions with the real system). 
Fig. \ref{fig:introfigure} shows the tile of trajectory tracking from start to end. 
The evaluation on convergence of M-DeMoRL was observed till 80 epochs and we concluded at reward gains of 2 times the observed performance with SAC baseline training. This clearly illustrates the accelerating nature of M-DeMoRL and justifies feasibility of training directly on the hardware setup.  

\textbf{Quadrupedal walking:}
We extend our approach to StochLite (shown in Figure \ref{fig:legandstochresults} right) for walking, which is a low-cost robot consisting of $12$ servos and an inertial measurement unit (IMU) to detect
the body pose. StochLite is also capable of measuring joint angles (hip and knee) and torques via encoders and motor current sensors respectively. The robot dynamics model consists of $6$ floating degrees-of-freedom and $12$ actuated degrees-of-freedom. The simulator for StochLite is PyBullet \cite{coumans2021}. The states consist of the torso pose, velocity (both translational and rotational components) and the slope orientation (roll and pitch). Slope orientations are obtained via the IMU orientation and the joint angle encoders.

We execute tracking end-foot trajectories via a low-level control law, while the higher level walking control policy focuses on yielding optimal trajectory parameters similar to \cite{Hwangboeaau5872, paigwar2020robust}. Those parameters form the actions/controls, to be obtained from the RL policy.
We use a well-defined trajectory generator, which is in turn shaped via a neural network based policy. This framework, when combined with our sample efficient M-DeMoRL algorithm, allows to train policies in much shorter time ($<60$ episodes of true environment interactions with 1000 interactions per episode). 
Figure~\ref{fig:legandstochresults} (right) shows the reactive behavior of policy in terms of the modulations applied by the neural network controller over the pre-defined elliptical trajectory. These modulations are able to shift the trajectories along X, Y and Z directions. The learned response validates that the neural policy to model the correspondence of pitch and roll with the X and Y shifts. As the pitch orientation increases, the quadruped tries to increase the X-shifts so that the leg moves forward and stabilizes the torso. Similarly, the increase in roll is accompanied by again an increase in the Y-shifts in the rolling direction. We also validate the policy with successful transfer to hardware. Previously, \cite{haarnoja2018learning} trained policies in Minitaur quadruped in 160 epochs using SAC. Due to increasing complexity, we will consider direct hardware training for quadrupeds in future.

%% file: Sections/Conclusion.tex
\section{Conclusion}
\label{sec:conclusion}
We have investigated the role of leveraging the model-based optimisation with online learning to accelerate model-free RL algorithms. 
With the emphasis to develop a real-time controller, this work primarily defines a generalised framework that could be used with the existing MPC algorithms and off the shelf Mf-RL algorithms to train 
efficiently. 
Simulation results show that our formulation yields sample efficient algorithms, as the underlying online learning tracks for the best policy benefiting the convergence of the Mf-RL policy. 
We also show that the training is 
faster than prior Mf-Mb methods. 
Finally, we show hardware-in-loop training with a 2R leg and we successfully generate hardware transferable policies for quadruped walking.
While preliminary hardware results are shown, we look forward to better gaits and more efficient direct training on hardware as future works. The video for our experiment can be found at: \href{https://stochlab.github.io/redirects/MDeMoRLPolicies.html}{stochlab.github.io/redirects/MDeMoRLPolicies.html}.


%% file: Sections/Proofs.tex
\newpage
\appendix
\section{Proofs of Lemma \ref{nlem1} and Theorem \ref{thm:regrettheorem}}
\label{sec:proofs}

Let us consider ${\mathcal{M}}$ and ${\mathcal{M}}_r$ as the approximated and real MDP with dynamics model $f_\phi$ and ${f}$ respectively. Let the total variation distance between them be bounded by $\epsilon_f$ (see \cite{NEURIPS2019_MBPO}). This dynamics model predicts both the next state distribution and rewards. 
The corresponding MPC objective is represented as $J$ and ${J}_r$ respectively. Here, ${J}$ denotes that the costs are calculated from the approximated reward function setting whereas ${J}_r$ is obtained from rollouts in the true MDP. Now, we will derive the bounds on the performance improvement in a similar way as demonstrated in \cite{NEURIPS2019_MBPO} and \cite{morgan2021model}, however with consideration and assumptions related to the convexity of the losses.

\begin{proof}[Proof of Lemma 1]
For any stochastic dynamics model $f$ and reward function $r$, considering the cost of a trajectory in an MDP with policy $\pi_\eta$ and value function $V_\zeta$ is given by,
\begin{equation}
    C\left(\mathbf{x_t}, \mathbf{u_t}\right) = \sum^{H-1}_{h = 0} \gamma^h c(x_{t,h},u_{t,h}) + \gamma^H c_H(x_{t,H}) 
\end{equation}
where, $\gamma$ is the discount factor, $c(x_{t,h},u_{t,h}) = - r(x_{t,h},u_{t,h})$ and $c_H$ is the terminal cost calculated as $-V_\zeta (x_{t,H})$. Let $c_{max}$ be the bound on this cost.

Now, to realize the maximum improvement in the approximated MDP while using the policy parameters ($\tilde{\eta}_{t}$), obtained from the shift model, we use a formulation motivated by the bound formulated in Lemma B.3 in \cite{NEURIPS2019_MBPO}. We consider $p_\phi$ as the discounted state-action visitation corresponding to $f_\phi$ (similarly $p$ for $f$) and superscript $h$ to resemble the notations of \cite{NEURIPS2019_MBPO}.

\begin{align*}
    {J} & \left( x_t, \tilde{\eta}_{t}\right)  - J_r\left(x_t, \tilde{\eta}_{t}\right) \\
    = & \; \mathbb{E}_{\mathbf{u_t} \sim \pi_{\tilde{\eta}_t},\mathbf{x_t}\sim f_\phi} \left[ \sum_{h=0}^H \gamma^h c(x_{t,h}, u_{t,h}) + \gamma^H c_H (x_{t,H}) \right] \\ 
    & - \mathbb{E}_{\mathbf{u_t} \sim \pi_{\tilde{\eta}_t}, \mathbf{x_t} \sim f} \left[ \sum_{h=0}^H \gamma^h c(x_{t,h}, u_{t,h}) + \gamma^H c_H(x_{t,H}) \right] \\
    = & \sum_{\mathbf{x_t}, \mathbf{u_t}} (p_\phi(x,u) - p(x,u)) \ c(x,u) \\
    \leq & \! \sum_{\mathbf{x_t}, \mathbf{u_t}} \! \sum_{h=0}^{H-1} \! \gamma^h \ (p^h_\phi(x_{t,h},u_{t,h}) \! - \! p^h(x_{t,h},u_{t,h})) \ c(x_{t,h},u_{t,h}) \\
    & + \gamma^H \ (p^H_\phi(x_{t,H},u_{t,H}) - p^H(x_{t,H},u_{t,H}))V_\zeta(x_{t,H}) \\
    \leq & 2 \ c_{max} \sum_{h=0}^{H-1} \gamma^h \ h \epsilon_f + \gamma^H 2 \ V_{max} H \epsilon_f \\
    = & 2 \ c_{max} \frac{(H-1)\gamma^{H+1} - H\gamma^H + \gamma}{(1-\gamma)^2} \ \epsilon_f + \gamma^H 2 \ V_{max} H \epsilon_f
\end{align*}
where, $|({p^h}(x,u) - p^h_\phi(x,u)) | \leq h \epsilon_f$ is inherited from Lemma B.2 in \cite{NEURIPS2019_MBPO}, the uncertainty in dynamics approximation. 
\end{proof}

\begin{proof}[Proof of Theorem 1]
From Lemma-1, we know that,
\begin{equation}
    {J}\left(\tilde{\eta}_{t}\right) \leq J_r\left(\tilde{\eta}_{t}\right) + R_{f,H} 
\end{equation}
and subtracting $J_r\left({\eta}_{t}^\star\right)$ from both sides of Eq~(4) results in
\begin{equation}
    {J}\left(\tilde{\eta}_{t}\right)-J_r\left({\eta}_{t}^\star\right) \leq J_r\left(\tilde{\eta}_{t}\right)-J_r\left(\eta_{t}^\star\right) + R_{f,H} 
\end{equation}
where LHS corresponds to the instantaneous regret incurred by rollouts on approximate MDP (with ${J}$) using shifted parameters ($\tilde{\eta}_t$) and on true MDP (with $J_r$) using the DMD-optimized parameters~(${\eta}_t$).

Now, to get the cumulative regret for $T$ decision steps, both sides of Eq~(5) should be summed over $T$ and can be shown as,
\begin{align}
    \sum_{t=0}^T \left({J}\left(\tilde{\eta}_{t}\right)-J_r\left({\eta}_{t}^\star\right) \right) \leq & \sum_{t=0}^T \left( J_r\left(\tilde{\eta}_{t}\right)-J_r\left(\eta_{t}^\star\right)\right) + \sum_{t=0}^T R_{f,H} \\
    {Re}_{T} \left(\boldsymbol{\eta}_{T}\right) \leq & \sum_{t=0}^T \left( J_r\left(\tilde{\eta}_{t}\right)-J_r\left(\eta_{t}^\star\right)\right) + T \ R_{f,H}
\end{align}
Based on \cite{hall2013dynamical}, the DMD update rule directly results in
\begin{equation}
    \sum_{t=0}^T \left( J\left(\tilde{\eta}_{t}\right)-J_r\left(\eta_{t}^\star\right)\right) \leq \frac{D_{\max }}{\alpha_{T+1}}+\frac{4 M}{\alpha_{T}} W_{\Phi_{t}}\left(\boldsymbol{\eta}_{T}\right)+\frac{G_{\ell}^{2}}{2 \sigma} \sum_{t=1}^{T} \alpha_{t} 
\end{equation}
Substituting Eq~(8) in Eq~(7), we finally get the bound on the maximum regret as
\begin{equation*}
    {Re}_{T} \left(\boldsymbol{\eta}_{T}\right) \leq \frac{D_{\max }}{\alpha_{T+1}}+\frac{4 M}{\alpha_{T}} W_{\Phi_{t}}\left(\boldsymbol{\eta}_{T}\right)+\frac{G_{\ell}^{2}}{2 \sigma} \sum_{t=1}^{T} \alpha_{t} + T \ R_{f,H},
\end{equation*}
which completes the proof.
\end{proof}






%% file: root.bbl
\begin{thebibliography}{10}

\bibitem{Rajeswaran-RSS-18}
A.~Rajeswaran, V.~Kumar, A.~Gupta, G.~Vezzani, J.~Schulman, E.~Todorov, and
  S.~Levine, ``Learning complex dexterous manipulation with deep reinforcement
  learning and demonstrations,'' in {\em Proceedings of Robotics: Science and
  Systems}, (Pittsburgh, Pennsylvania), June 2018.

\bibitem{peng2020learning}
X.~B. Peng, E.~Coumans, T.~Zhang, T.-W.~E. Lee, J.~Tan, and S.~Levine,
  ``Learning agile robotic locomotion skills by imitating animals,'' in {\em
  Robotics: Science and Systems}, 07 2020.

\bibitem{lee2020learning}
J.~Lee, J.~Hwangbo, L.~Wellhausen, V.~Koltun, and M.~Hutter, ``Learning
  quadrupedal locomotion over challenging terrain,'' {\em Science robotics},
  vol.~5, no.~47, 2020.

\bibitem{Levine13GPS}
S.~Levine and V.~Koltun, ``Guided policy search,'' in {\em Proceedings of the
  30th International Conference on Machine Learning} (S.~Dasgupta and
  D.~McAllester, eds.), vol.~28 of {\em Proceedings of Machine Learning
  Research}, (Atlanta, Georgia, USA), pp.~1--9, PMLR, 17--19 Jun 2013.

\bibitem{nagabandi2018neural}
A.~Nagabandi, G.~Kahn, R.~S. Fearing, and S.~Levine, ``Neural network dynamics
  for model-based deep reinforcement learning with model-free fine-tuning,'' in
  {\em 2018 IEEE International Conference on Robotics and Automation (ICRA)},
  pp.~7559--7566, IEEE, 2018.

\bibitem{nagabandi2020deep}
A.~Nagabandi, K.~Konolige, S.~Levine, and V.~Kumar, ``Deep dynamics models for
  learning dexterous manipulation,'' in {\em Conference on Robot Learning},
  pp.~1101--1112, PMLR, 2020.

\bibitem{NEURIPS2018_handfultrials}
K.~Chua, R.~Calandra, R.~McAllister, and S.~Levine, ``Deep reinforcement
  learning in a handful of trials using probabilistic dynamics models,'' in
  {\em Advances in Neural Information Processing Systems} (S.~Bengio,
  H.~Wallach, H.~Larochelle, K.~Grauman, N.~Cesa-Bianchi, and R.~Garnett,
  eds.), vol.~31, Curran Associates, Inc., 2018.

\bibitem{pourchot2018cemrl}
Pourchot and Sigaud, ``{CEM}-{RL}: Combining evolutionary and gradient-based
  methods for policy search,'' in {\em International Conference on Learning
  Representations}, 2019.

\bibitem{Williams2017imppi}
G.~Williams, N.~Wagener, B.~Goldfain, P.~Drews, J.~M. Rehg, B.~Boots, and E.~A.
  Theodorou, ``Information theoretic mpc for model-based reinforcement
  learning,'' in {\em 2017 IEEE International Conference on Robotics and
  Automation (ICRA)}, pp.~1714--1721, 2017.

\bibitem{NEURIPS2019_MBPO}
M.~Janner, J.~Fu, M.~Zhang, and S.~Levine, ``When to trust your model:
  Model-based policy optimization,'' in {\em Advances in Neural Information
  Processing Systems} (H.~Wallach, H.~Larochelle, A.~Beygelzimer,
  F.~d\textquotesingle Alch\'{e}-Buc, E.~Fox, and R.~Garnett, eds.), vol.~32,
  Curran Associates, Inc., 2019.

\bibitem{lowrey2018plan}
K.~Lowrey, A.~Rajeswaran, S.~Kakade, E.~Todorov, and I.~Mordatch, ``Plan
  online, learn offline: Efficient learning and exploration via model-based
  control,'' in {\em International Conference on Learning Representations},
  2019.

\bibitem{morgan2021model}
A.~S. Morgan, D.~Nandha, G.~Chalvatzaki, C.~D'Eramo, A.~M. Dollar, and
  J.~Peters, ``Model predictive actor-critic: Accelerating robot skill
  acquisition with deep reinforcement learning,'' {\em arXiv preprint
  arXiv:2103.13842}, 2021.

\bibitem{wagener2019online}
N.~Wagener, C.~an~Cheng, J.~Sacks, and B.~Boots, ``An online learning approach
  to model predictive control,'' in {\em Proceedings of Robotics: Science and
  Systems}, (FreiburgimBreisgau, Germany), June 2019.

\bibitem{haarnoja2018soft}
T.~Haarnoja, A.~Zhou, P.~Abbeel, and S.~Levine, ``Soft actor-critic: Off-policy
  maximum entropy deep reinforcement learning with a stochastic actor,'' in
  {\em International Conference on Machine Learning}, pp.~1861--1870, PMLR,
  2018.

\bibitem{NEURIPS2019_deepdynamicmodel}
J.~Z. Kolter and G.~Manek, ``Learning stable deep dynamics models,'' in {\em
  Advances in Neural Information Processing Systems} (H.~Wallach,
  H.~Larochelle, A.~Beygelzimer, F.~d\textquotesingle Alch\'{e}-Buc, E.~Fox,
  and R.~Garnett, eds.), vol.~32, Curran Associates, Inc., 2019.

\bibitem{lillicrap2015continuous}
T.~P. Lillicrap, J.~J. Hunt, A.~Pritzel, N.~Heess, T.~Erez, Y.~Tassa,
  D.~Silver, and D.~Wierstra, ``Continuous control with deep reinforcement
  learning.,'' in {\em ICLR (Poster)}, 2016.

\bibitem{hall2013dynamical}
E.~Hall and R.~Willett, ``Dynamical models and tracking regret in online convex
  programming,'' in {\em International Conference on Machine Learning},
  pp.~579--587, PMLR, 2013.

\bibitem{todorov2012mujoco}
E.~Todorov, T.~Erez, and Y.~Tassa, ``Mujoco: A physics engine for model-based
  control,'' in {\em 2012 IEEE/RSJ International Conference on Intelligent
  Robots and Systems}, pp.~5026--5033, IEEE, 2012.

\bibitem{coumans2021}
E.~Coumans and Y.~Bai, ``Pybullet, a python module for physics simulation for
  games, robotics and machine learning.'' \url{http://pybullet.org},
  2016--2021.

\bibitem{Hwangboeaau5872}
J.~Hwangbo, J.~Lee, A.~Dosovitskiy, D.~Bellicoso, V.~Tsounis, V.~Koltun, and
  M.~Hutter, ``Learning agile and dynamic motor skills for legged robots,''
  {\em Science Robotics}, vol.~4, no.~26, 2019.

\bibitem{paigwar2020robust}
K.~Paigwar, L.~Krishna, S.~Tirumala, N.~Khetan, A.~Sagi, A.~Joglekar,
  S.~Bhatnagar, A.~Ghosal, B.~Amrutur, and S.~Kolathaya, ``Robust quadrupedal
  locomotion on sloped terrains: A linear policy approach,'' 2020.

\bibitem{haarnoja2018learning}
T.~Haarnoja, S.~Ha, A.~Zhou, J.~Tan, G.~Tucker, and S.~Levine, ``Learning to
  walk via deep reinforcement learning,'' {\em arXiv preprint
  arXiv:1812.11103}, 2018.

\end{thebibliography}
